\documentclass[sigconf,nonacm]{acmart}
\AtBeginDocument{%
  }

\usepackage{algorithmic}
\usepackage{algorithm}
\usepackage[ruled,vlined,algo2e]{algorithm2e}
\usepackage{setspace}
\usepackage{multirow}
\usepackage{cleveref}
\usepackage{enumitem}
\usepackage{threeparttable}
\usepackage{graphicx}
\usepackage{caption}
\usepackage{subcaption}
\usepackage{adjustbox} 
\usepackage{makecell}
\usepackage{booktabs}
\usepackage{fontawesome}

\renewcommand\footnotetextcopyrightpermission[1]{} 
\acmDOI{} 
\acmISBN{} 
\begin{document}

\title{AIF: Asynchronous Inference Framework for Cost-Effective Pre-Ranking}
\author{Zhi Kou}
\email{kouzhi.kz@alibaba-inc.com}
\author{Xiang-Rong Sheng}
\email{xiangrong.sxr@alibaba-inc.com}
\affiliation{%
  \institution{Taobao \& Tmall Group of Alibaba}
  \city{Beijing}
  \country{China}
}

\author{Shuguang Han}
\email{shuguang.sh@alibaba-inc.com}
\author{Zhishan Zhao}
\email{zhaozhishan.zzs@alibaba-inc.com}
\author{Yueyao Cheng}
\email{yueyao.cyy@alibaba-inc.com}
\affiliation{%
  \institution{Taobao \& Tmall Group of Alibaba}
  \city{Beijing}
  \country{China}
}

\author{Han Zhu}
\email{zhuhan.zh@alibaba-inc.com}
\author{Jian Xu}
\email{xiyu.xj@alibaba-inc.com}
\author{Bo Zheng}\authornote{Corresponding author.} 
\email{bozheng@alibaba-inc.com}
\affiliation{%
  \institution{Taobao \& Tmall Group of Alibaba}
  \city{Beijing}
  \country{China}
}

\renewcommand{\shortauthors}{Zhi Kou et al.}
\newcommand{\sheng}[1]{\textcolor{red}{#1}}

\begin{abstract}
In industrial recommendation systems, pre-ranking models based on deep neural networks (DNNs) commonly adopt a sequential execution framework:
feature fetching and model forward computation are triggered \textit{only after} receiving candidates from the upstream retrieval stage. This design introduces inherent bottlenecks, including redundant computations of identical users/items and increased latency due to strictly sequential operations, which jointly constrain the model's capacity and system efficiency.

To address these limitations, we propose the Asynchronous Inference Framework (\textbf{AIF}), a cost-effective computational architecture that decouples \textit{interaction-independent} components---those operating within a single user or item---from real-time prediction. AIF reorganizes the model inference process by performing user-side computations in parallel with the retrieval stage and conducting item-side computations in a nearline manner. This means that interaction-independent components are calculated just once and completed before the real-time prediction phase of the pre-ranking stage. As a result, AIF enhances computational efficiency and reduces latency, freeing up resources to significantly improve the feature set and model architecture of interaction-independent components. Moreover, we delve into model design within the AIF framework, employing approximated methods for interaction-dependent components in online real-time predictions. By co-designing both the framework and the model, our solution achieves notable performance gains without significantly increasing computational and latency costs.  This has enabled the successful deployment of AIF in the Taobao display advertising system.

\end{abstract}






\maketitle

\section{Introduction}

\begin{figure}[!t] 
  \centering
  \includegraphics[width=\linewidth]{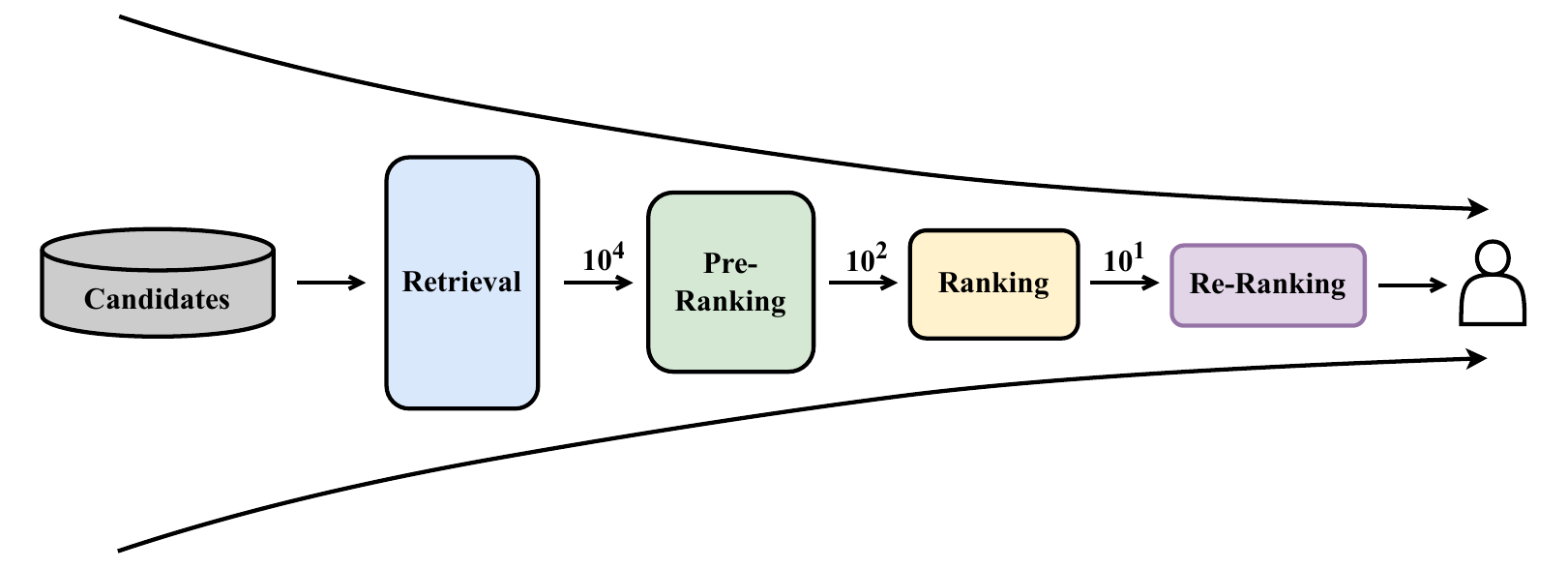} 
  \caption{An illustration of the typical multi-stage cascade
architecture for industrial recommender systems.}
  \label{fig:multi-stage}
\end{figure}

Industrial recommendation systems often employ a multi-stage cascade architecture for progressive candidate selection, aiming to balance the overall effectiveness with resource costs. As shown in ~\Cref{fig:multi-stage}, candidate items progress through multiple stages, including candidate retrieval~\cite{zhu2018tdm, HuangSSXZPPOY2020FacebookEBR}, pre-ranking~\cite{WangZJZZGCold, ma2021towards}, ranking~\cite{cheng2016wide, zhou2018din, MaZYCHC2018MMOE}, and re-ranking~\cite{PeiZZSLSWJGOP2019PersonalizedRerank}, to ultimately produce personalized recommendation results. Serving as the bridge between the retrieval and ranking stages, pre-ranking is crucial for selecting the most valuable subset from an initial candidate set of approximate $10^4$  items (from retrieval) to around $10^2$ high-quality candidates, directly influencing the quality of input for subsequent ranking stages.

\begin{figure*}[t]
    \begin{subfigure}[b]{0.4\textwidth}
        \centering
        \adjustbox{valign=c}{\includegraphics[scale = 0.45]{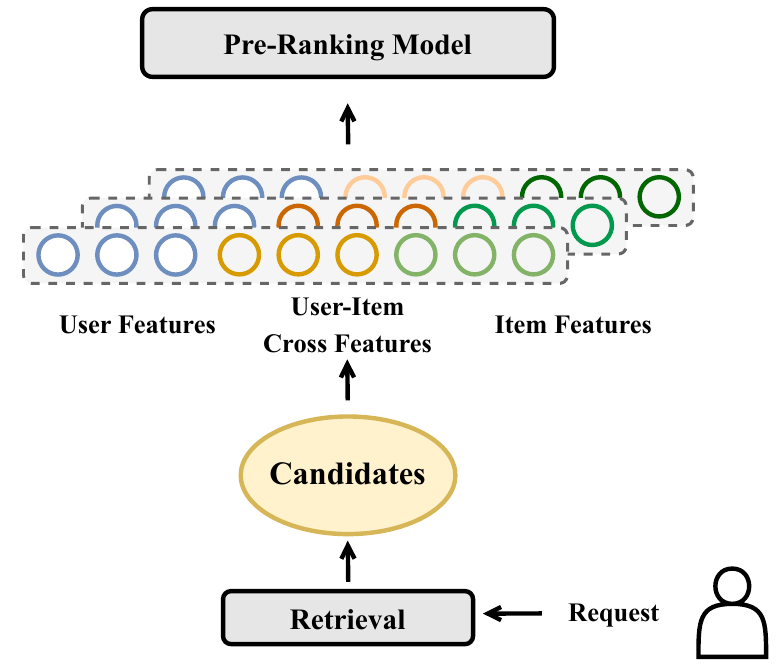}}
        \caption{Typical Sequential Inference Pipeline.}
        \label{fig:typical}
    \end{subfigure}%
    ~
    \begin{subfigure}[b]{0.6\textwidth}
        \centering
        \adjustbox{valign=c}{\includegraphics[scale = 0.45]{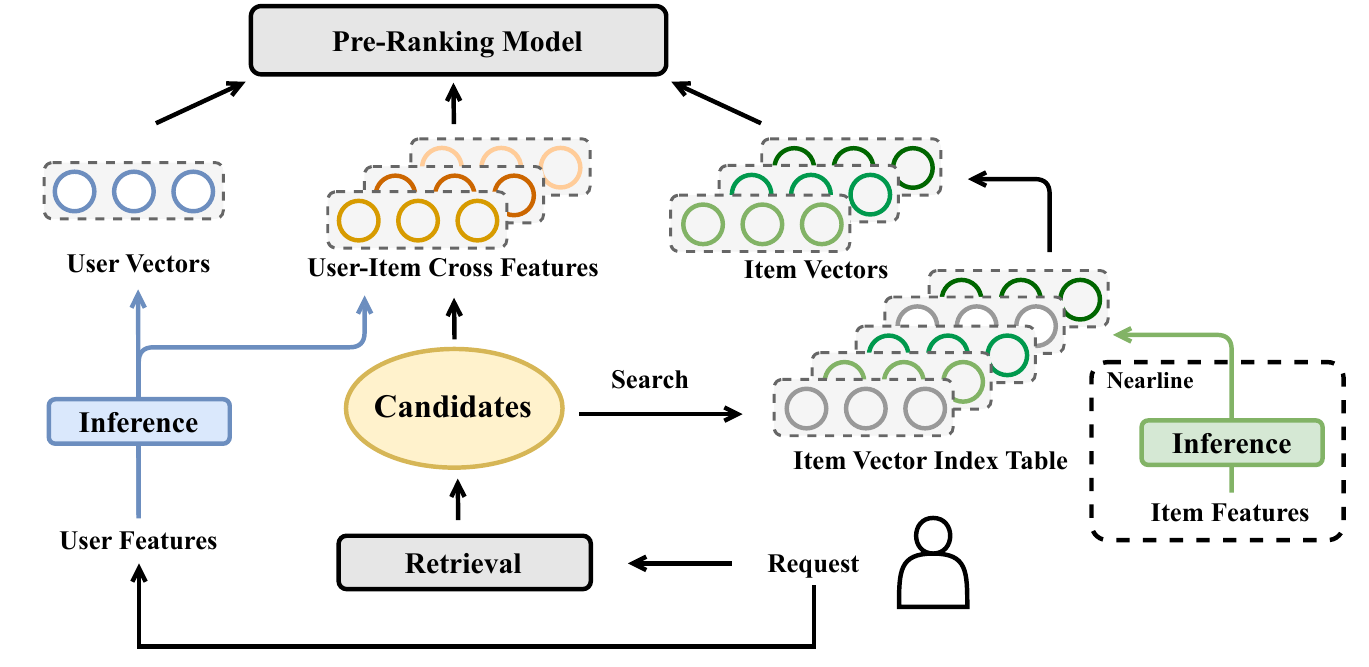}}
        \caption{Asynchronous Inference Framework.}
        \label{fig:aif}
    \end{subfigure}
    \caption{A comparison between sequential inference framework and AIF. In AIF, the intra-computation of user-side features and the user-side component of cross features(the blue part) are performed through online asynchronous inference, while the intra-computation of item-side features(the green part) utilizes nearline asynchronous inference.}  
    \label{fig:aif_compare}    
\end{figure*}

The pre-ranking stage faces two primary constraints: 1) real-time prediction adheres to strict latency constraints and 2) pre-ranking needs to approximate the ranking accuracy of the subsequent ranking stage but with restricted computational complexity.  To address these challenges, industrial systems typically employ a ``divide-and-conquer + simplification'' strategy. Once the retrieval stage provides the candidate set, the system partitions it into mini-batches (e.g., 1,000 items per batch) for separate and parallel model inference to optimize inference latency. To minimize the computation overhead, each mini-batch is processed using a lightweight model~\cite{WangZJZZGCold}.

As illustrated in ~\Cref{fig:aif_compare} (a), a typical pre-ranking model inference process consists of two sequential phases: 1) \textbf{feature fetching} and 2) \textbf{model forward computation}. In the first phase, the model accesses user and item features from feature storage systems, along with performing operations such as cross-feature generation. Following this, it constructs model input tensors by indexing the model embedding matrices based on feature values. In the second phase, the system performs model forward computation for each pair of (user, item) using a shallow multilayer perceptron.

However, pre-ranking models operating within this typical sequential execution framework have inherent disadvantages:

\begin{itemize}[leftmargin=*]

\item \textbf{Redundant Computation}: Within the same user request, user-side computations, including fetching user features (e.g. behavior sequences) and performing internal user computation (like self-attention operations on these sequences~\citep{SunLWPLOJ2019Bert4Rec}), are repeated across multiple mini-batches. Meanwhile, item-side computations are redundantly processed across multiple requests, since the same item may be recommended for multiple users.

\item \textbf{Increased Latency}: Candidate retrieval, feature fetching, and model forward computation of the pre-ranking stage create a feedforward chain, yet not all processes are sequentially dependent. For instance,  user-side computations can be decoupled from other parts of inference and performed in parallel with the retrieval stage. Yet, the typical sequential framework, which requires all features to be fetched before proceeding to model forwarding, leads to increased latency.

\item \textbf{Restricted Model Capacity}: Under the current framework, pre-ranking models often forego complex cross-features and sophisticated model structures, leading to a performance gap compared to the ultimate ranking model.

\end{itemize}

To address these issues, we present a novel cost-effective asynchronous inference paradigm termed the \textbf{A}synchronous \textbf{I}nference \textbf{F}ramework (\textbf{AIF}). 
AIF is motivated by the insight that interaction-independent components, those operating within a single user or item, can be decoupled from the sequential pipeline and precomputed asynchronously, which eliminates the need for real-time computation and reduces redundant computation. Building on this insight, the AIF framework transforms the traditional sequential inference paradigm into an asynchronous computation framework. As shown in ~\Cref{fig:aif_compare} (b), AIF performs the asynchronous inference for interaction-independent components through the following strategies:
\begin{itemize}[leftmargin=*]
\item \textbf{Asynchronous user-side computation}: AIF asynchronously pre-executes user-side feature fetching and internal user network forward computation during candidate retrieval, thereby reducing the latency incurred during pre-ranking (the retrieval latency can now be used for pre-ranking). Additionally, since user-side internal network computation is executed only once, this approach significantly reduces computational costs associated with redundant operations across different mini-batches.

\item \textbf{Asynchronous item-side computation}: Similarly, for item-side features, AIF performs feature fetching and internal item network forward computation, storing pre-computed tensors for all items in a near-line manner. Here, near-line refers to a hybrid inference approach that performs offline computation while incorporating real-time updates to models or features (i.e. the computation is triggered once the model checkpoint or item feature is updated). It combines the real-time capability of online inference with the scalability and efficiency of offline inference. This allows item-side computation to be completed in advance and executed only once, thus reducing computational overhead.

\begin{table*}[h]
\centering
\caption{Comparison of asynchronous inference at different stages. More \faStar indicates higher overhead/performance.}
\label{tab:stage_comparison}
\resizebox{\textwidth}{!}
{
\begin{tabular}{l|c|c|c|c}
\toprule
\textbf{Method} & \textbf{Computation Overhead}  & \textbf{Storage Overhead} 
 &\textbf{Latency Overhead} &  \textbf{Timeliness}\\
\midrule
Offline Asynchronous  Inference    & \faStar \faStarO \faStarO \faStarO    &  \faStar \faStar \faStar \faStarO   & \faStarO \faStarO \faStarO \faStarO  & \faStar \faStarO  \faStarO \faStarO\\ 
Nearline Asynchronous  Inference     &  \faStar \faStar \faStarO \faStarO    &  \faStar \faStar \faStar \faStar   & \faStarO \faStarO \faStarO \faStarO & \faStar \faStar \faStar  \faStarO\\ 
Online Asynchronous  Inference   &  \faStar \faStar \faStar  \faStarO   &  \faStar \faStar \faStarO  \faStarO  & \faStar \faStarO \faStarO \faStarO  & \faStar \faStar \faStar \faStar \\ 
\midrule
Real-time Inference & \faStar \faStar \faStar \faStar   & \faStar \faStarO  \faStarO \faStarO  & \faStar \faStar \faStar \faStar    & \faStar \faStar \faStar \faStar \\ 
\bottomrule
\end{tabular}
}
\end{table*}

\item \textbf{Asynchronous cross features processing acceleration}: Cross features are typically hard to be pre-executed asynchronously. To this end, we divide the computation of cross features into two parts for acceleration: pre-processing of user components, and joint processing of user and item components to generate cross features. In AIF, the first part, i.e. the user-side component, is executed asynchronously to minimize latency.
\end{itemize}

With these strategies, AIF improves both computational efficiency and reduces latency, allowing the use of saved resources
to significantly scale the interaction-independent model component.

It is important to note that while AIF can optimize non-interaction computations, such as user-side or item-side intra-computation,  the user-item interaction components cannot be fully pre-processed asynchronously. This limitation arises because the candidate set is only determined immediately before the pre-ranking stage, which cannot be accessed before the retrieval stage or nearline. Therefore, we also investigate how to apply approximate methods within the AIF framework to manage user-item interactions, seeking to augment the model capacity with acceptable computational and latency costs. To this end, AIF introduces an approximate interaction computation between asynchronous inferred user-side and item-side tensors, enriching the feature interaction capability. Meanwhile, AIF also employs an efficient long-term user sequence modeling technique based on multi-modal representations.

By decoupling interaction-independent computations (handled asynchronously via AIF) from interaction-dependent computations (approximated for efficiency), we propose a comprehensive cost-effective strategy to optimize pre-ranking models. This approach not only enhances performance in large-scale recommendation systems but also introduces novel methodologies for future research. 
The AIF framework has been successfully deployed in Taobao's display advertising, achieving a +8.72\% increase in CTR (click-through rate) and a +5.80\% boost in RPM (Revenue Per Mille) with no added latency and computational costs rising by less than 15\%.

\section{Asynchronous Inference Framework}
\label{sec:aif}
Model inference computations can be categorized by their reliance on user-item feature interactions. Interaction-independent computations, those that operate entirely within a single user or item, can be decoupled and executed asynchronously. In contrast, interaction-dependent computations require both user and item data, typically requiring real-time processing. In this work, we propose AIF: a framework that applies asynchronous inference to interaction-independent computations and employs approximation techniques for interaction-dependent ones.

\textbf{Implementation of AIF.} Pre-computation by asynchronous inference involves generating and caching intermediate results prior to the online pre-ranking phase, thereby reducing computational overhead and latency during real-time predictions. In pre-ranking models, asynchronous inference can be applied across three stages: offline, nearline, and online(at retrieval). \Cref{tab:stage_comparison} presents a comparative analysis of the trade-offs between these strategies. Based on this evaluation and the distinct characteristics of user and item features, our framework employs online asynchronous inference for user-side tasks and nearline asynchronous inference for item-side tasks, guided by the following rationale:
\begin{itemize}[leftmargin=*]
\item {User-side computations.} User features (e.g., user behaviors) are updated frequency and requires real-time online computation. Note that user features are independent of the candidate set and can be accessed upon request arrival, prior to the retrieval stage. To leverage this, AIF adopts an online asynchronous inference approach, executing a user-side network in parallel with the retrieval stage.

\item {Item-side computations.} Item features depend on the post-retrieval candidate set, making online parallel computation infeasible. However, their low update frequency allows cross-request reuse of precomputed features. This inherent trade-off between freshness and efficiency motivates AIF's nearline asynchronous inference design for item-side processing.
\end{itemize}

\textbf{Model Design under AIF.} 

Unlike conventional models that rely solely on raw information for feature fetching and model forwarding, models within the AIF framework employ hybrid inputs that encompass both raw information and precomputed user-side and item-side vectors at the online pre-ranking stage. While the asynchronous inference architecture efficiently reduces overhead for non-interactive components and enables the pre-ranking model to incorporate additional features, user-item interaction computations cannot be fully asynchronized due to their essential dependence on real-time candidate sets.
To address this challenge, we systematically explore approximation methods for interaction modeling within the AIF framework. Specifically, we propose the following two approaches:
\begin{enumerate}[leftmargin=*]
\item An efficient dimensionality expansion of asynchronously inferred vectors via approximated feature interactions.
\item An efficient similarity-based long-term user behavior modeling via Locality Sensitive Hashing technique with multi-modal representations.
\end{enumerate}
Through the co-design of the framework and model, our solution achieves significant performance gains without considerable increases in computational and latency costs.

\section{Implementation of AIF}
\label{sec:fe}
In this section, we will delve into the implementation details of AIF.

\subsection{Asynchronous Online Inference for User-Side Computations}

\begin{figure}[h]
    \centering
    \includegraphics[width=\columnwidth]{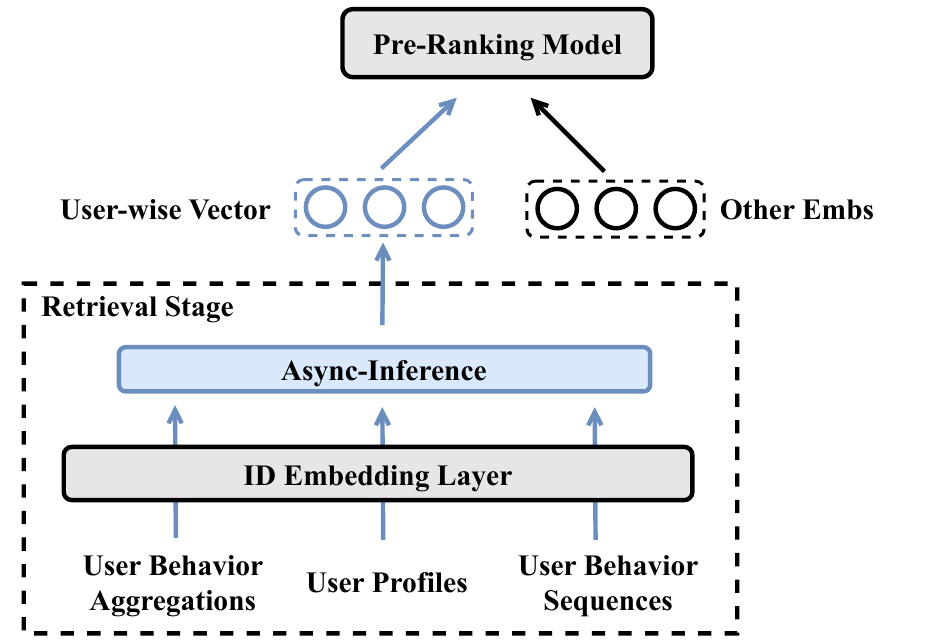}
    \caption{Asynchronous online inference for user-side computations.}
    \label{fig:async1}
\end{figure}

In online asynchronous inference, the system's central coordinator (\textit{Merger}), which integrates outputs from modules to produce final personalized recommendations, interacts with the real-time prediction platform (\textit{RTP}) twice: 1) online asynchronous inference for user-side pre-computations, parallelized with upstream candidate retrieval (see ~\Cref{fig:async1}) and 2) real-time prediction during the pre-ranking phase to compute final scores.

\textbf{Pre-Computation Phase.}  During retrieval, the Merger concurrently triggers RTP to execute online asynchronous inference. It first fetches user features, including user profiles and behavior sequence  from feature storage system and indexes the model embedding, $\mathbf{U}_\text{profile}$ and $\mathbf{U}_\text{seq} $, based on feature values.

\label{sec:user_feature}

As for the model structure, we adopts self-attention on behavior sequences and cross-attention computation between profiles $\mathbf{U}_\text{profile} \in \mathbb{R}^{1 \times d^U}$ and behavior sequence $\mathbf{U}_\text{seq} \in \mathbb{R}^{l \times d^U}$. Specifically, AIF first projects the user profile and behavior sequence embeddings into the same dimensionality:
\begin{equation}
\begin{aligned}
    \hat{\mathbf{U}}_\text{profile} &= \mathbf{U}_\text{profile}\mathbf{W}_\text{profile}^\top \in \mathbb{R}^{1 \times d} \\
    \hat{\mathbf{U}}_\text{seq} &= \mathbf{U}_\text{seq}\mathbf{W}_\text{seq}^\top  \in \mathbb{R}^{l \times d},
\end{aligned}
\end{equation}
where $l$ is the length of behavior sequence.
Then  we apply self-attention on behavior sequences:
\begin{equation}
    \mathbf{U}_\text{self\_attention} = \text{Pooling}(\text{FFN}(\text{Softmax}(\frac{ \hat{\mathbf{U}}_\text{seq} \hat{\mathbf{U}}_\text{seq} ^\top}{\sqrt{d}} ) \hat{\mathbf{U}}_\text{seq})) \in \mathbb{R} ^ {1 \times d}, \\ 
\end{equation}
and cross-attention computation between profiles and behavior sequences:
\begin{equation}
    \mathbf{U}_\text{profile\_attention} = \text{Softmax}(\frac{ \hat{\mathbf{U}}_\text{profile} \hat{\mathbf{U}}_\text{seq} ^\top}{\sqrt{d}} ) \hat{\mathbf{U}}_\text{seq} \in \mathbb{R} ^ {1 \times d}. 
\end{equation}
The self-attention and cross attention mechanism effectively captures complex relationships between user profiles and behavior sequences.
After the computation, the combined user vector is cached immediately. Note that this user-level processing avoids mini-batch redundancy and introduces no latency to real-time pre-ranking.

\textbf{Real-Time Prediction Phase.} At pre-ranking, Merger initiates a second RTP request incorporating cached user vector. The RTP combines the precomputed vector with raw data to compute final pre-ranking scores, maintaining low latency.

\subsection{Asynchronous Nearline Inference for Item-Side Computations}
\begin{figure}[h] 
  \centering
  \includegraphics[width=\columnwidth]{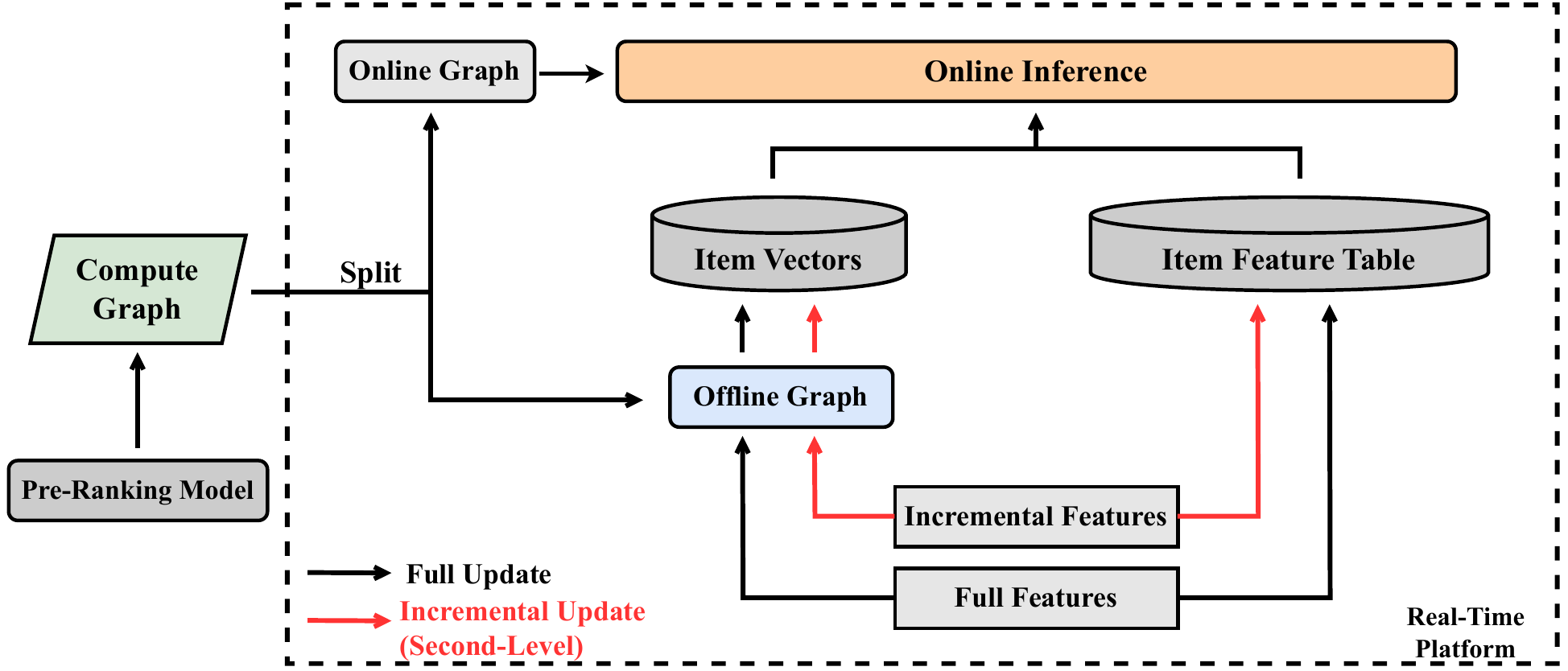}
  \caption{Asynchronous nearline inference for
item-side computations.}
  \label{fig:n2o}
\end{figure}

While online async-computation for items is impractical (due to dependency on post-retrieval candidate sets and high computational cost caused by redundant cross-request computations), AIF employs nearline asynchronous inference for item-side computations, decoupled from real-time requests and triggered by model/feature updates (\Cref{fig:n2o}). Specifically, AIF computes item vectors through:
\begin{itemize}[leftmargin=*]
\item \textbf{Dimensionality reduction}: For an item with concatenated embeddings $\mathbf{I}\in \mathbb{R}^{1 \times d^I}$, a multi-layer perceptron projects it to low-dimensional vector:
\begin{equation}
\hat{\mathbf{I}} = \text{MLP}(\mathbf{I}) \in \mathbb{R}^{1 \times d}
\end{equation}
\item \textbf{Update-triggered execution}: The above-mentioned computation is initiated upon model parameter updates or item feature changes, generating vectors for the full candidate set stored in an indexing table.
\end{itemize}

At pre-ranking, the precomputed item vectors are combined with precomputed user vectors to compute final pre-ranking scores.

From the effectiveness perspective, the nearline asynchronous inference is feasible because item features (e.g., static attributes) exhibit lower time sensitivity than user behavior sequences, enabling nearline processing without performance degradation. From the efficiency perspective, the nearline inference eliminates redundant cross-request computations and bypasses online latency constraints (unlike asynchronous online inference that needs adhere to the latency constraint of the preceeding candidate retrieval).

\subsection{Online Pre-Caching for Cross Features}
Traditional user-item cross-features are generated entirely during pre-ranking, creating latency bottlenecks. AIF decouples the process of cross features into user-side pre-processing phase and real-time generation phase:
\begin{itemize}[leftmargin=*]
\item \textbf{User-Side Pre-Processing}: The user component is executed asynchronously in parallel to candidate retrieval via online asynchronous inference.
\item \textbf{Real-Time Generation}: The cross feature generation is completed at pre-ranking using candidate item information and the pre-processed user component.
\end{itemize}
\begin{figure}[h] 
  \centering
  \includegraphics[width=\columnwidth]{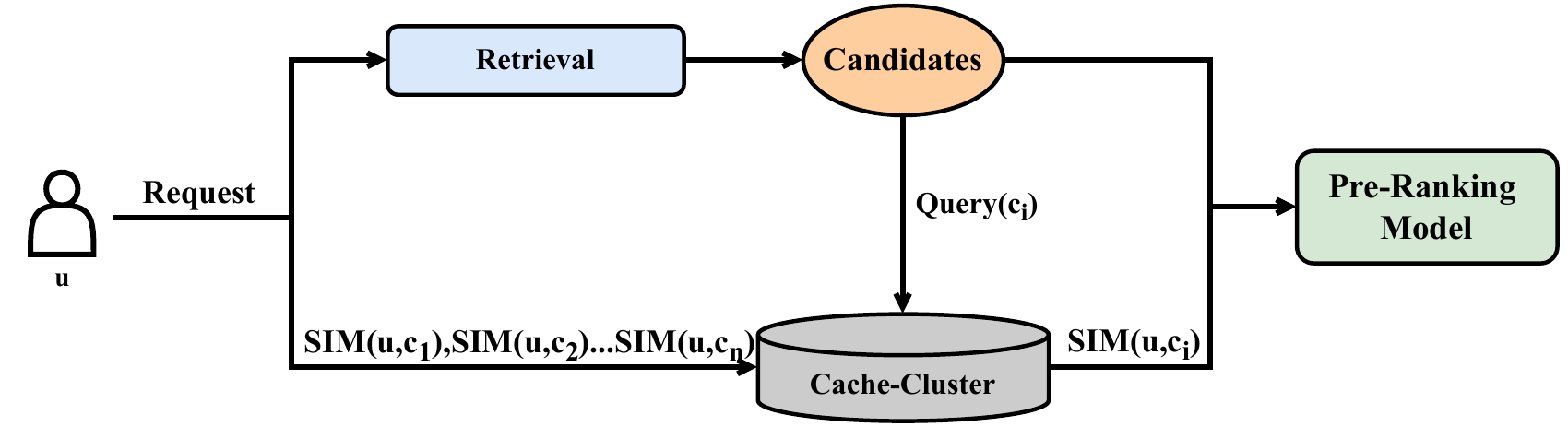}
  \caption{Illustration of pre-caching mechanism for SIM features.}
  \label{fig:sim}
\end{figure}

Here, we illustrate this process with the SIM-hard feature~\cite{PiZZWRFZG2020SIM}. SIM-hard is the long term user behavior sequence that employed in many industrial systems (length$\sim10^5$). Specifically, SIM-hard preprocesses the long-term user sequence offline as <user, category, sub\_sequence>. 
During pre-ranking, subsequences are selected based on the candidate items' category, which cause a dependency on post-retrieval candidate category information.

In practice, SIM-hard faces latency bottlenecks due to remote feature access and parsing processes.  To address the latency issues of SIM-hard and other cross-features, AIF introduces a feature pre-caching mechanism. Parallel to retrieval, AIF pre-caches parsed subsequences for all possible user-category combinations of the requesting user using an LRU cache cluster (~\Cref{fig:sim}). During pre-ranking, AIF directly indexes relevant subsequences from the cache cluster, eliminating online fetching and parsing delays.

\subsection{Engineering Effort for AIF}
Deploying AIF requires addressing model version consistency challenges caused by distributing computations across multiple stages. 

\textbf{Online Asynchronous Inference.} To synchronize user-side computations between asynchronous inference and pre-ranking, AIF employs a unique hashed key, consisting of the request ID and user nickname, for each request to implement consistent hashing. This approach ensures the consistency of user-side features used by asynchronous inference and the pre-ranking model. Furthermore, AIF adopts an Arena memory pool for the high-frequency updates and caching of user-side features and user-side component of cross features, thereby significantly enhancing the efficiency of feature access and processing.

\textbf{Nearline Asynchronous Inference.} To ensure item-side consistency between nearline asynchronous inference and pre-ranking, AIF employs an index table for N2O that supports both full and incremental updates, similar to the item feature index table. Consequently, the N2O result index table is updated synchronously whenever the original item feature index table undergoes full or incremental updates, guaranteeing feature consistency between nearline asynchronous inference and the pre-ranking model. The construction of the index table is based on offline high-priority CPU resources, utilizing highly concurrent processes for parallel computation to ensure the timeliness of the N2O results.

\section{Model Design under AIF}
\label{sec:fi}
The user-item feature interaction is crucial for recommendation models. Although the asynchronous inference enhances the pre-ranking model's capacity by scaling the interaction-independent structures, a challenge remains in enhancing interaction-dependent structures within the AIF framework. This section focuses primarily on two types of interaction computation under AIF: 1) efficient expansion of asynchronously inferred vectors, and 2) efficient modeling of target item and user behavior sequences, i.e., efficient user behavior modeling.

\subsection{Efficient Expansion of Asynchronously Inferred Vectors}
During the online real-time prediction phase, AIF processes hybrid inputs comprising precomputed user/item-side vectors from asynchronous inference and real-time fetched embeddings. To further enhance the effectiveness of asynchronously inferred vectors in AIF, the most straightforward approach is to expand the dimensionality of the async-inferred vectors. However, these vectors are subjected to substantial compression compared to their original features (over 10x for user-side features and over 3x for item-side features) to maintain optimal system performance. As a result, directly expanding the dimensionality of asynchronously inferred vectors would impose additional computational overhead on the pre-ranking model. Therefore, we propose the Bridge Embedding Approximation (\textbf{BEA}) methods, which enables efficient computation for asynchronously inferred user/item-side vectors with higher dimensionality.

\textbf{Bridge Embedding Approximation.} Inspired by the Poly-Encoder architecture~\cite{humeau2020poly}, we introduces $n$ learnable $d$-dimensionality bridge embeddings $\mathbf{B}=[\mathbf{b}_0,\mathbf{b}_1,\cdots,\mathbf{b}_{n-1}]^\top \in \mathbb{R}^{n\times d}$. These embeddings are utilized to compute cross-attention with $m$ groups of user-side feature embeddings $\mathbf{U}\in \mathbb{R}^{m\times d}$ and item-side feature embeddings $\mathbf{I} \in \mathbb{R}^{b\times d}$. For user-side, we generates $n$ async-inferred vectors instead of a single vector using the user-side asynchronous inference network $f(\cdot|\Theta_u)$. For item-side, we generates $n$ attention weights that are computed through the N2O pipeline. In the pre-ranking stage, AIF performs a weighted summation of the $n$ user-side vectors and the item-side attention weights.

\begin{algorithm}[h]
\caption{Bridge Embedding Approximation.}
\label{alg:BEA}
\renewcommand{\algorithmicrequire}{\textbf{Input:}}
\renewcommand{\algorithmicensure}{\textbf{Output:}}
\setstretch{1.2}
\begin{algorithmic}[1]
    \REQUIRE $\mathbf{B} \in \mathbb{R}^{n\times d}, \mathbf{U}\in \mathbb{R}^{m\times d}, \mathbf{I} \in \mathbb{R}^{b\times d}, f(\cdot|\Theta_u)$
    \ENSURE $\hat{\mathbf{v}}\in \mathbb{R}^{b \times d'}$.

    \STATE $\mathbf{W} = \text{Softmax}( \frac{ \mathbf{B} \mathbf{U}^\top}{\sqrt d})\in \mathbb{R}^{n \times m}$ \tcp*{Cross-attention between the bridge and user-side ID embeddings}

    \STATE $\mathbf{V}=f(\mathbf{U},\mathbf{W}|\Theta_u) \in \mathbb{R}^{n\times d'}$
    \tcp*{Inference user-side vectors with weighted features}

    \STATE $\hat{\mathbf{w}}=\text{Softmax}(\frac{\mathbf{I}\mathbf{B}^\top}{\sqrt d})\in \mathbb{R}^{b \times n}$ \tcp*{Cross-attention between the bridge and item-side ID embeddings}

    \STATE $\hat{\mathbf{v}} = \hat{\mathbf{w}} \mathbf{V}\in \mathbb{R}^{b \times d'}$
    \tcp*{Weighted sum}

    \RETURN $\hat{\mathbf{v}}$
\end{algorithmic}
\end{algorithm}

Here, original user features are used to perform pairwise interactions with $n$ learnable randomly initialized bridge embeddings, rather than with item features. Since $n$ is much smaller than the size of the item candidate set, this approach significantly increases the dimensionality of user-side asynchronously inferred vectors with an efficient feature interactions method. These bridge embeddings are trained end-to-end alongside the pre-ranking model using online learning and remain fixed during inference. Additionally, because the bridge embeddings are decoupled from item candidates, the primary computations for feature interactions, cross attention mechanism, can be performed asynchronously online (for user-side) and nearline (for item-side), further reducing online inference latency. The implementation details are illustrated in the ~\Cref{alg:BEA}.

\subsection{Efficient User Behavior Modeling}
\label{sec:ub}

User behavior modeling, capturing latent interests from historical interactions, is critical for recommendation models~\cite{zhou2018din,zhou2019dien,ShengYGWCZCZG0J2024EnhancingTaobaoMM,ChangZFZGLHLNSG2023TWIN}. Common approaches to user behavior modeling involve computing similarities between the target item and historically interacted items, and then perform pooling of historical item embeddings, guided by the computed similarities. Note that while asynchronous inference enables long-term behavior sequence integration (~\Cref{sec:fe}), the computational complexity of the dot product similarity calculations $O(bld)$ poses prohibitive costs, where $b\sim10^4$, $l\sim10^5$ and $d\sim10^2 $ represent the inference batch size, behavior sequence length, and embedding dimensionality, respectively.  This high computational cost poses a significant burden in current long term user behavior modeling approaches, hindering their deployment with large-scale recommendation systems.


\textbf{Approximation of the Similarity Computation.} To address this challenge, we propose a Locality Sensitive Hashing (LSH)-based approximation method, reducing embedding dimensionality while preserving similarity relationship. The main idea of LSH is that if two high-dimensional representations are similar (i.e., possess minimal Euclidean distance) in the original representation space, their corresponding LSH hash transformations will produce identical hash values with high probability. For simplicity, we adopt a 2-bit LSH hash transformation for each item multi-modal embedding $\mathbf{M}_i \in \mathbb{R}^{1 \times d}$:
\begin{equation}
\begin{aligned}
&\mathbf{M}_{\text{hash}_i} = \text{Relu}(\text{Sign}(\mathbf{M}_i\mathbf{W}_\text{hash}^\top) \in \{0,1\}^{1 \times d'},
\end{aligned}
\end{equation}
where  $\mathbf{W}_\text{hash} \in \mathbb{R}^{d' \times d}$ is sampled from the standard normal distribution $\mathcal{N}(0,1)$ and is shared across all embeddings. 

The decision to use multi-modal embeddings for similarity computation is driven by the reason that multi-modal embeddings are pre-trained and static, eliminating the need for parameter updates in the pre-ranking model~\cite{ShengYGWCZCZG0J2024EnhancingTaobaoMM}. This allows for pre-processing without the concern of version control issues between offline training and online serving.

The similarity computation between such binary signatures can be efficiently implemented through the sum of bit-wise XNOR (inverse of the XOR($\oplus$), return 1 when both inputs are the same, otherwise return 0) results:
\begin{equation}
\mathbf{M}_{\text{sim}_{ij}}=\frac{\overline{ \mathbf{M}_{\text{hash}_i} \oplus \mathbf{M}_{\text{hash}_j}}}{d'}  =\frac{1}{d'}\sum_{k=1}^{d'}\overline{ \mathbf{M}_{\text{hash}_i^{(k)}} \oplus \mathbf{M}_{\text{hash}_j^{(k)}} },
\end{equation}
where $\mathbf{M}_{\text{hash}_i}$ and $\mathbf{M}_{\text{hash}_j}$ are two hashed item multi-modal embeddings, and $\mathbf{M}_{\text{hash}^{(k)}}$ indicates the $k_{th}$ bit of $\mathbf{M}_{\text{hash}}$.

Binary signatures enable efficient encoding through decimal representation. For example, an 8-bit binary number $00110101_{(2)}$ can be precisely represented as the decimal number $53_{(10)}$. The similarity calculation can be transformed into a combination of XNOR and PopulationCount operations (counting the number of $1$s in the binary representation of an integer), enabling lossless compression of the LSH-based binary signatures. Leveraging on LSH technique, AIF is able to introduce longer user behavior sequences with multi-modal representations without increasing the computation or latency cost. 

In practical implementation scenarios, the hashed multi-modal representations are pre-computed offline over the entire item candidate set, resulting an uint8 embedding index table. Moreover, given that uint8 has only 256 possible values, the PopulationCount operation can be replaced with a lookup operation in a $1\times 256$-dimensional embedding table, further accelerating similarity computation. 

\textbf{Update Methods.} The LSH signatures for existing items are computed once and remain static, leveraging the strong generalization capability of pre-trained multi-modal representations~\cite{ShengYGWCZCZG0J2024EnhancingTaobaoMM}. Updates to LSH signatures are applied exclusively to new items, ensuring computational efficiency. To maintain real-time effectiveness for new items, we employ an incremental message queue that dynamically processes updates, enabling seamless integration of new entries without recalculating existing signatures.

\textbf{User Behavior Modeling.} Given the pairwise similarities between the target items $\mathbf{M}_{\text{item}} \in \{0,1\}^{b \times d'}$ and historically interacted items $\mathbf{M}_{\text{seq}} \in \{0,1\}^{l \times d'}$:
\begin{equation}
    \mathbf{M}_{\text{sim}}=\text{LSH-Similarity}(\mathbf{M}_{\text{item}},\mathbf{M}_{\text{seq}}) \in \mathbb{R}^{b \times l},
\end{equation}
we adopt them to two behavior modeling techniques: 
DIN~\cite{zhou2018din} and SimTier~\cite{ShengYGWCZCZG0J2024EnhancingTaobaoMM}.
\begin{itemize}[leftmargin=*]
    \item \textbf{DIN:} Given the historical item embedding sequence
    $\mathbf{U}_{\text{seq}} \in \mathbb{R}^{l \times d}$, 
    DIN produces a weighted sum of historical item embeddings:
	\begin{equation}
	\text{DIN}(\mathbf{U}_{\text{seq}}, \mathbf{M}_{\text{sim}}) = \mathbf{M}_{\text{sim}}(\mathbf{U}_{\text{seq}},\mathbf{W}_{\text{seq}}^\top) \in \mathbb{R}^{b \times d}
	\end{equation}
    \item \textbf{SimTier:} SimTier discretizes similarity scores into $N$ predefined tiers. For each tier corresponding to a specific similarity interval, SimTier counts the number of scores within that range, generating an $N$-dimensional histogram-like vector
	\begin{equation}
    	\mathbf{h}_{\text{tier}} = \text{Concat}(\text{Count}_1, \text{Count}_2, ..., \text{Count}_N)
	\end{equation}
	where $\text{Count}_i$ indicates the number of similarity scores in the $i$-th tier.
\end{itemize} 

\section{Experiments}

We conducted a comprehensive evaluation of AIF, analyzing both model performance and system performance in Taobao's display advertising system.

\subsection{Experimental Settings}

\textbf{Production Dataset}. The dataset used in the experiments consists of 8 days of impression and ranking logs collected from Taobao's online display advertising system. These logs amount to billions of user-item interactions. The first week of logs is used for training, while the last day is reserved for evaluation.

\textbf{Training Details}. AIF was conducted on 32 H20 GPUs with Adam optimizer and distributed data parallelism for 1 epoch. The mini-batch size is set to 5,000 samples per GPU, with a learning rate of 1e-4 and weight decay of 1e-5. The user-side features contain user attributes, contextual attributes, user behavior sequences and aggregated features. The item-side features contain item attributes and multi-modal representations. The loss function used in AIF is $\Delta$NDCG-Based Pari-wise Rank Alignment Loss used in COPR~\cite{zhao2023copr}:
\begin{equation}
    L=\sum_{i<j}\Delta NDCG(i,j)\log[1+e^{-(\frac{y_i* bid_i}{y_j*bid_j}-1)}].
\end{equation}

\textbf{Compared Methods}. AIF is compared against a baseline that adopts the COLD~\cite{WangZJZZGCold} architecture with multi-modal representations and COPR~\cite{zhao2023copr} Loss Function. To further analyze the performance of AIF, we also conducted ablation studies on different asynchronous inference techniques. Each model is trained in one epoch~\cite{ZhangSZJHDZ2022OneEpoch,zong2025recis}.

\textbf{Metrics}. We adopt two groups of metrics for evaluation.
\begin{itemize}[leftmargin=*]
    \item \textbf{Model Performance Metrics}. We evaluate model performance using both offline and online metrics. The \textbf{offline metrics} include HitRatio of TopK (HR@K) and Group-AUC (GAUC), and the top 10 candidates selected by the ranking model are treated as relevant items. The \textbf{online metrics} include click-through rate (CTR) and revenue per mille (RPM), which respectively correspond to user experience and platform revenue. Results were obtained through a large-scale A/B test comparing the baseline model and the AIF model. Traffic was randomly divided via a hash of user identity keys (user ID), ensuring equitable distribution between control and treatment groups (50/50 split) over a 14-day period. This methodology guarantees consistent user assignment to avoid cross-contamination and enables statistically significant comparisons of model performance.
    \item \textbf{System Performance Metrics}. To evaluate the impact of AIF on system performance, we introduce system performance metrics, including average response time (avgRT), the 99th percentile response time (p99RT), and maximum Queries Per Second (maxQPS). Generally, higher QPS and lower RT indicate lower computational resource requirements for a given model. We also compared the demand for extra storage across different methods.
\end{itemize}

\textbf{Significance Tests}. Online results are assessed using bootstrapping with 1000 resamples (95\% confidence intervals). Metric improvements, such as CTR and RPM, are considered statistically significant if their confidence intervals do not cross zero. The online performance gains were validated as significant under this test, with all intervals firmly excluding zero.

\subsection{Evaluation on Model Performance}
\label{sec:model_performance}

\subsubsection{Effectiveness of the Asynchronous Feature Enhancement}

As shown in ~\Cref{tab:model_performance}, the "upper bound" of the base COLD model can be achieved by directly using all features as input into the model (+7.83pt in GAUC), termed \textbf{Base(full features)}. However, it leads to significant increases in online computation and latency, making it impractical to deploy in the pre-ranking stage. By introducing asynchronous feature enhancement and approximated feature interactions in the pre-ranking model, AIF delivers performance close to it (+7.29pt in GAUC), achieving significant improvement both in HR@100 and GAUC compared to the current online version of the pre-ranking model, termed \textbf{Base}. We also conducted ablation studies on different components of AIF. Experiments demonstrate that each component of AIF contributes to a measurable performance improvement in HR@100 and GAUC. It is important to note that AIF expands user-side features by up to 7x and item-side features by up to 3x through asynchronous feature enhancement compared to the current online version.

\begin{table}[h]
\centering
\caption{Model performance comparison of asynchronous feature enhancement.}
\label{tab:model_performance}
\resizebox{\columnwidth}{!}
{
\begin{tabular}{l|cc|cc}
\toprule
\textbf{Method} & \textbf{HR@100} & \textbf{GAUC} & \textbf{CTR} & \textbf{RPM} \\
\midrule
Base            & -          & -        &- &-              \\ 
Base(full features)    &  +8.45pt          & +7.83pt  & - & -                       \\ 
AIF        & \textbf{+7.91pt}          & \textbf{+7.29pt}   &\textbf{+8.72\%}    & \textbf{+5.80\%}                      \\
\midrule
AIF w/o Async-Vectors       & +3.99pt          & +3.71pt       & +4.43\%         & +3.36\%               \\ 
AIF w/o Pre-Caching SIM       & +5.97pt          & +6.13pt        & +6.11\%         & +4.79\%              \\
AIF w/o BEA    & +5.86pt    & +6.09pt  & +7.19\%    & +4.02\%      \\
AIF w/o Long-term User Behavior          & +5.43pt    & +5.98pt    & +6.45\%   & +3.71\%           \\ 
\midrule
Base with +15\% candidates & - & - & +3.75\% & +1.69\% \\
Base with +15\% parameters & - & - & +1.96\% & +1.07\% \\

\bottomrule
\end{tabular}
}
\end{table}


To evaluate the overall performance of AIF in a production environment, we conducted a large-scale online A/B test using real traffic, where the proposed methods were deployed to serve real users and advertisers. As shown in \cref{tab:model_performance}, AIF achieves significant improvements, with +8.72\% in CTR and +5.80\% in RPM, demonstrating its effectiveness in real-world scenarios. 

\subsubsection{Effectiveness of the Bridge Embedding Approximation}

\begin{figure}[h] 
  \centering
  \includegraphics[width=\linewidth]{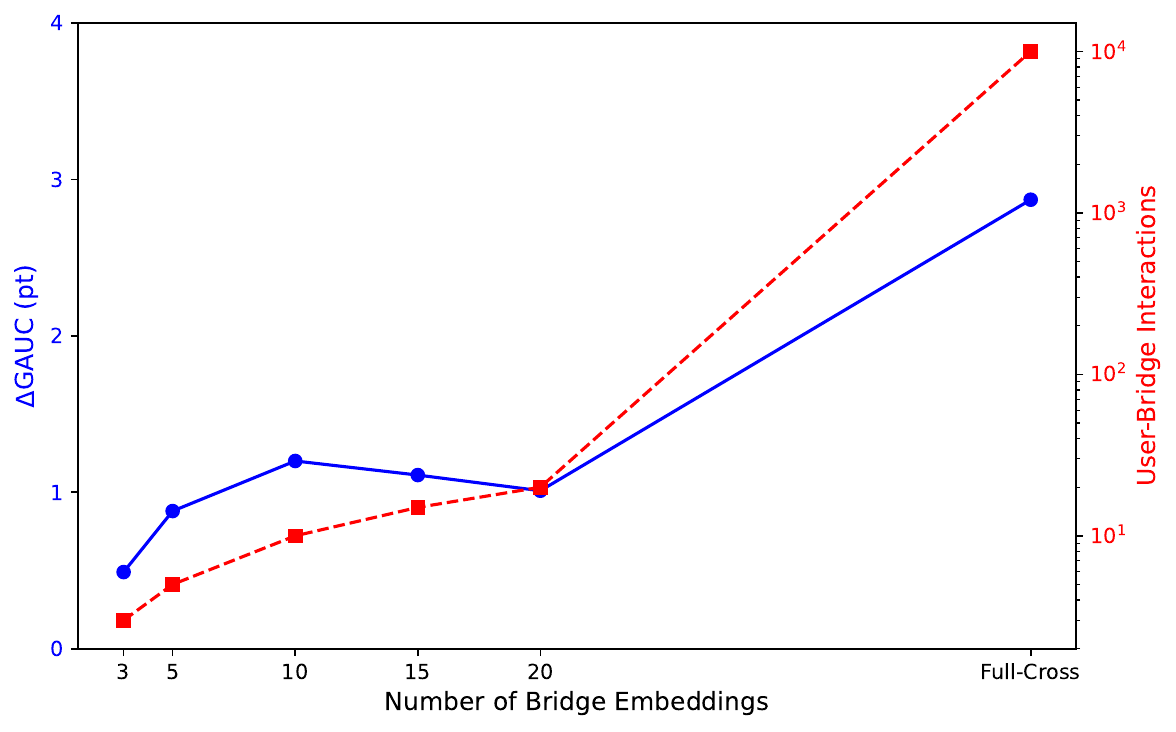} 
  \caption{An ablation study on number of bridge embeddings in BEA.}
  \label{fig:bea}
\end{figure}

As shown in ~\Cref{tab:model_performance}, by introducing Bridge Embedding Approximation, AIF achieves an increase of 2.05pt in HR@100 and 1.20pt in GAUC. As for the online A/B test in real traffic, the BEA approach achieves improvements of up to +1.53\% in CTR and +1.78\% in RPM. We conducted an ablation study on the impact of the number of bridge embeddings on model performance and computational cost. As shown in \Cref{fig:bea}, model performance improves incrementally as the number of bridge embeddings increases. However, beyond a threshold (e.g., 10), performance plateaus or declines due to training divergence caused by over parameterization (blue line). Similarly, the interaction between the user-side asynchronous features and the bridge features introduced by BEA also increases with the number (red line).

We also compare the perfomance-complexity trade-off between BEA and full interaction with candidate items as bridge embeddings (directly computing the full interactions between all candidate items and user-side asynchronous features), termed Full-Cross. Since the candidate set typically contains more than $10^4$ items, the computational complexity of Full-Cross can reach hundreds of times that of BEA, making it difficult to apply in online systems.

\subsubsection{Effectiveness of the Approximated Long-Term User Behavior Modeling}
AIF also achieves a significant increase in HR@100 and GAUC by introducing longer user behavior sequences with less computation complexity via LSH. To evaluate the effectiveness of approximated long-term user behavior modeling in AIF, we conducted ablation studies on various combinations as discussed in \Cref{sec:ub}. Due to the uint8 8x-compression for the original binary signatures in LSH ($d_{id}=d_{mm}=8d_{lsh}$), by replacing the ID-based attention module in DIN or the multi-modal-based similarity module in SimTier with multi-modal LSH similarity mechanism, AIF achieves a reduction in computational complexity of attentions and similarities by 43.75\%. Furthermore, AIF benefits from reusing computation results of LSH-similarity when applied in both modules, achieving significant reduction in the computational complexity of the attention and similarity modules up to 93.75\% compared to the original DIN and SimTier frameworks with only 0.45pt descent in GAUC. In addition to significantly reducing the theoretical computational complexity, longer user behavior sequences modeling via LSH in AIF also leads to resource reduction in terms of online feature storage, read,  and memory access/copying overhead.

\begin{table}[h]
\centering
\caption{Model Performance and Complexity Comparison of Efficient Long-Term User Behavior Modeling.}
\label{tab:lsh_performance}
\resizebox{\columnwidth}{!}{
\begin{tabular}{l|c|cc}
\toprule
\textbf{Method} & \textbf{GAUC} & \textbf{Complexity} & \textbf{Reduction}\\
\midrule
DIN + SimTier     &    -   &  $bl(d_{id} + d_{mm})$  & -   \\ 
LSH-DIN + SimTier &     -0.28pt        &      $bl(d_{lsh} + d_{mm})$  & -43.75\%  \\
DIN + LSH-SimTier &     -0.37pt        &      $bl(d_{id} + d_{lsh})$  & -43.75\% \\
MM-DIN + SimTier &     -0.23pt        &      $bld_{mm}$ & -50\% \\
\midrule
LSH-DIN + LSH-SimTer(\textbf{AIF}) &  -0.45pt  & $bld_{lsh}$ & -93.75\%\\
\bottomrule
\end{tabular}
}
\end{table}

\subsubsection{Effectiveness of Asynchronous Inference Framework}
To further highlight the advantages of AIF under the same resource and latency constraints, we compared its online performance with two baseline approaches: 1) directly expanding the candidate item set of the pre-ranking stage by 15\% and 2) expanding the parameters of the pre-ranking network by 15\%. As illustrated in \Cref{tab:model_performance}, under the constraints of no latency increase and computational cost increase of less than 15\%, AIF significantly outperforms both the candidate item set expansion and network parameter expansion approaches in terms of CTR and RPM, showcasing its superior efficiency and effectiveness. It means that performance improvements mainly stem from the asynchronous inference framework, which decouples feature computation and model inference. By prefetching user/item features in \Cref{sec:fe} and enabling complex interaction modeling in \Cref{sec:fi}, AIF allows the pre-ranking model to utilize:
\begin{itemize}
    \item \textbf{Richer feature sets}: Historical user behavior, contextual signals, and cross-modal item embeddings.
    \item \textbf{Advanced architectures}: Attention-based interaction layers and multi-task learning, which were previously infeasible under strict latency budgets. 
\end{itemize}
This architectural shift prioritizes feature quality and model capacity within fixed computational constraints, directly addressing the pre-ranking stage’s bottleneck of oversimplified models. Due to its impressive performance, AIF has been successfully deployed to serve the main traffic of Taobao's display advertising system in the pre-ranking stage since October 2023.

\subsection{Evaluation on System Performance}

As a cost-effective model architecture, AIF significantly enhances model performance while simultaneously optimizing system performance. We conducted the same A/B tests described in ~\Cref{sec:model_performance} to validate the system performance advantages of different parts of AIF. As shown in ~\Cref{tab:system_performance}, the use of LSH for long-term user behavior modeling and the pre-caching mechanism for SIM features in AIF achieve significant reductions in RT and increases in QPS compared to directly incorporating these features into the model, resulting in lower computational costs for the pre-ranking model. In addition, the introduction of Async-Vectors and BEA only results in a slight system performance degradation, which also remains within an acceptable range.



\begin{table}[h]
\centering
\caption{System performance comparison of different methods.}
\label{tab:system_performance}
\resizebox{\columnwidth}{!}{%
\begin{tabular}{l|cc|c|c}
\toprule
\textbf{Method} & \textbf{avgRT} & \textbf{p99RT} & \textbf{maxQPS} & \textbf{\makecell{Extra \\ Storage}} \\

\midrule
Base            & -                  & -           & -     & -   \\
\midrule
+ Async-Vectors     & +0.02\%   & +0.05\%   & -0.75\%   & $\checkmark$  \\
+ SIM               & +30.43\%  & +18.02\%  & -19.23\%  & $\times$  \\
\quad + Pre-Caching & +0.05\%   & +0.03\%   & -0.52\%   & $\checkmark$  \\
+ BEA               & +0.25\%   & +0.50\%   & -1.27\%   & $\checkmark$  \\
+ Long-term User Behavior       & +45.12\%  & +51.80\%  & -46.47\%  &$\times$     \\
\quad + LSH         & +0.03\%   & +0.09\%   & -4.58\%   & $\times$      \\
\midrule
\textbf{AIF}        & +0.31\%   &  +0.55\%  & -6.72\%   & $\checkmark$     \\ 
\bottomrule
\end{tabular}
}
\end{table}

As shown in ~\Cref{sec:model_performance}, the complexity of AIF deployment is mainly due to three additional storage requirements: the N2O index table for item-side async-vectors and BEA, the transmission of user-side async-vectors, and the online pre-caching memory pool. The N2O index table is designed to store only the final item-side async-vectors, making it significantly smaller compared to the original item index table. To minimize transmission overhead, user-side asynchronous vectors are encoded using Base64. Although the memory pool size for online pre-caching is approximately 2-3 times the volume of actual requests, the application of LSH (Locality-Sensitive Hashing) significantly reduces the transmission and memory access associated with user behavior sequences. As a result, the actual deployment does not lead to a substantial increase in resource usage.

\section{Related Work}
The pre-ranking stage acts as an intermediary between the retrieval and ranking stages and is often seen as a simplified version of the ranking stage. Research aimed at improving pre-ranking models can be categorized into two areas: 1) enhancing the training objective to improve consistency between pre-ranking and ranking stages~\cite{qin2022rankflow,tang2018ranking,zhao2023copr}, and 2) increasing the model capacity of pre-ranking models~\cite{WangZJZZGCold,ma2021towards}.

This study focuses on enhancing pre-ranking model capacity in a cost-effective manner. Early efforts in developing pre-ranking models employed vector-product-based approaches~\cite{HuangHGDAH2013DSSM,li2022inttower}. These methods compute the pre-ranking score via inner products of user and item embeddings from a two-tower neural network.  To satisfy stringent latency requirements in real-world applications, these embedding vectors are generally  asynchronously pre-computed for efficient online inference.  Despite their computational efficiency, vector-product-based models lack user-item cross-features and interaction computations, which have been demonstrated to be highly effective in ranking models~\cite{guo2017deepfm,lian2018xdeepfm,WangFFW2017DCN,zhou2018din,BianWRPZXSZCMLX2022CAN,ShengZZDDLYLZDZ2021STAR,ShengGCYHDJXZ2023JRC}.

To address these limitations, recent industrial pre-ranking systems have transitioned to deep neural network (DNN)-based models, which introduce two key characteristics:  \textbf{enhanced interaction modeling} and \textbf{sequential online execution}. Concretely, cross-features and user-item interaction operations (e.g., attention mechanisms) are incorporated to improve expressiveness, albeit at the cost of higher computational overhead. Besides, all computations are performed online  in a sequential pipeline, eliminating asynchronous pre-computation.  For example,  \citet{WangZJZZGCold} employed a Multi-Layer Perceptron (MLP) architecture with cross features and integrated the SE (Squeeze-and-Excitation) block~\cite{hu2018squeeze} for feature selection. Similarly, \citet{ma2021towards} adopted a DNN-based model with feature selection to reduce the computational overhead. While these methods improve model capacity, their sequential execution introduces bottlenecks: redundant computations for identical users/items and increased latency due to strictly sequential operations. 

In this study,  we propose the Asynchronous Inference Framework (AIF), which combines the efficiency of vector-product-based asynchronous pre-computation with the capacity of DNN-based  models. AIF decouples interaction-independent components from the sequential pipeline, enabling asynchronous inference to eliminate redundancy. Simultaneously, interaction-dependent components are approximated via lightweight online modules, augmenting the model capacity with acceptable computational and latency costs.

\section{Conclusion and Discussion}

In this work, we present AIF, an asynchronous inference paradigm tailored to enhance pre-ranking models in a cost-effective manner for industrial systems. By simultaneously optimizing both the inference framework and model architecture, AIF achieves substantial performance improvements while maintaining low computational and latency overheads. Extensive offline and online experiments demonstrate AIF's superiority in both model and system performance. AIF has been successfully deployed in the Taobao display advertising system, showing practical viability for high-throughput industrial pre-ranking.

Importantly, AIF’s design principles extend beyond just the pre-ranking stage. The asynchronous inference methods enable more complex user-side computation or user feature prefetching with less latency. This also enables the ranking model to have more computational resources to incorporate other complex structures and features. When applied to Taobao's ranking models, the framework also achieves significant reductions in latency and computational resource usage. We believe that the strategies we developed in this study will serve as a valuable resource for practitioners looking to implement asynchronous inference paradigms in their systems.




\bibliographystyle{ACM-Reference-Format}
\bibliography{arxiv}

\end{document}